\documentclass[runningheads]{llncs}
\usepackage{graphicx}
\usepackage{amsmath,amssymb} 
\usepackage{multirow}
\usepackage{color}
\usepackage{booktabs}
\begin{document}
\title{ESNet: An Efficient Symmetric Network for Real-time Semantic Segmentation}
%
%
\author{
Yu Wang\inst{} \and
Quan Zhou\inst{}\thanks{Corresponding author: Quan Zhou} \and
Xiaofu Wu\inst{}
}
\titlerunning{ESNet for Real-time Semantic Segmentation} 

\authorrunning{Y. Wang et al.
}
%
\institute{
National Engineering Research Center of Communications and Networking, Nanjing University of Posts and Telecommunications, Nanjing 210003, China 
\email{wangy314159@163.com, quan.zhou@njupt.edu.cn, xfuwu@njupt.edu.cn}\\
}
\maketitle              
\begin{abstract}

The recent years have witnessed great advances for semantic segmentation using deep convolutional neural networks (DCNNs). However, a large number of convolutional layers and feature channels lead to semantic segmentation as a computationally heavy task, which is disadvantage to the scenario with limited resources. In this paper, we design an efficient symmetric network, called (\emph{ESNet}), to address this problem. The whole network has nearly symmetric architecture, which is mainly composed of a series of factorized convolution unit (FCU) and its parallel counterparts. On one hand, the FCU adopts a widely-used 1D factorized convolution in residual layers. On the other hand, the parallel version employs a transform-split-transform-merge strategy in the designment of residual module, where the split branch adopts dilated convolutions with different rate to enlarge receptive field. Our model has nearly 1.6M parameters, and is able to be performed over 62 FPS on a single GTX 1080Ti GPU. The experiments demonstrate that our approach achieves state-of-the-art results in terms of speed and accuracy trade-off for real-time semantic segmentation on CityScapes dataset.

\end{abstract}
%
%
%

\section{Introduction}

Semantic segmentation plays a significant role in image understanding \cite{krizhevsky2012imagenet,he2016deep,girshick2014rich}. From the perspective of computer vision, the task here is to assign a semantic label for each image pixel, which thus can be also considered as a dense prediction problem. Unlike conventional approaches that handle this challenge task by designing hand-craft features, deep convolutional neural networks (DCNNs) have shown their impressive capabilities in terms of end-to-end segmentation with full image resolution. The first prominent work in this field is fully convolutional networks (FCNs)\cite{long2017fully}, which are composed by a series of convolutional and max-pooling layers. After that, vast number of FCN-based network architectures \cite{Chen2016deeplab,zhao2017pyramid,xiao2017not} have been proposed and the remarkable progress have been achieved within segmentation accuracy. However, multiple stages of spatial pooling and convolution stride significantly reduce the dimension of feature representation, thereby losing much of the finer image structure. In order to address this problem, a more deeper architecture, named encoder-decoder network \cite{Badrinarayanan2015Segnet,Guosheng2017RefineNet,noh2015learning}, has become a trend, where the encoder network is utilized to abstract image features and the decoder counterpart is employed to sequentially recover image details. In the designment of network architecture, the residual network (ResNet) \cite{he2016deep} has been commonly adopted in recent years, where the residual layer allows to stack large amounts of convolutional layers, leading to the great improvement for both image classification \cite{krizhevsky2012imagenet,he2016deep,szegedy2015going} and semantic segmentation \cite{chao2017large,exploring2018lin,cong2019can}.

\begin{figure*}[!t]
\centerline{\includegraphics[width = 1.0\textwidth]{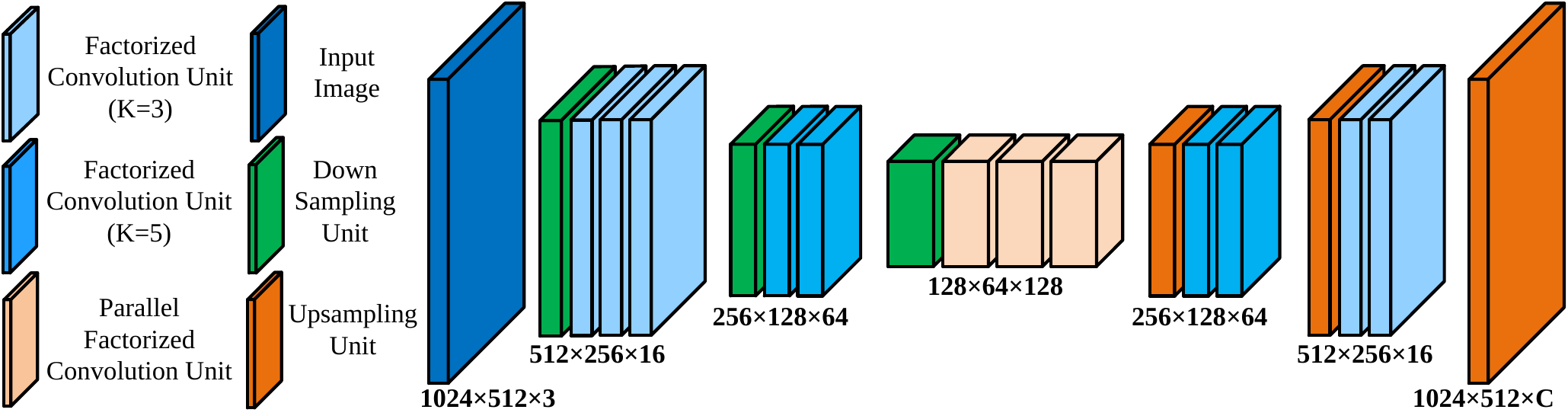}}
\caption{Overall symmetric architecture of the proposed ESNet. The entire network is composed by four components: down-sampling unit, upsampling unit, factorized convolution unit and its parallel version. (Best viewed in color)} \label{fig:Overview}
\end{figure*}

In spite of achieving impressive results, these accurate DCNNs neglect the implementing efficiency, which is a significant factor in limited resources scenarios. Considering running DCNNs on the mobile platforms (e.g., drones, robots, and smartphones), the designed networks are not only required to perform reliably (stability), but also required to conduct fast (real-time), suitable for embedded devices with space and memory constraints (compactness), and have low power consumption due to limited energy overhead (energy-saving). With this in mind, some preliminary research work \cite{wu2018cgnet,Treml2016speeding,Paszke2016enet,Mehta2018espnet} have been proposed to design lightweight networks that aim to develop efficient architectures for real-time semantic segmentation. However, these approaches usually focus on accelerating inference speed by aggressively reducing network parameters, which highly detriments segmentation performance. Therefore, pursuing the best performance with a good trade-off between accuracy and efficiency still remains an open research issue for the task of real-time semantic segmentation.

In this paper, we design a novel lightweight network called ESNet, adopting a nearly symmetric encoder-decoder architecture to address above problems. As shown in Fig. \ref{fig:Overview}, our ESNet is based on ResNet \cite{he2016deep}, which consists of four basic components, including down-sampling unit, upsampling unit, factorized convolution unit (FCU) and its parallel version. The core element of our architecture is parallel factorized convolution unit (PFCU), where a novel transform-split-transform-merge strategy is employed in the designment of residual layer, approaching the representational power of large and dense layers, but at a considerably lower computational complexity. More specifically, the PFCU leverages the identity mappings and multi-path factorized convolutions with 1D filter kernels. While the identity mappings allow the convolutions to learn residual functions that facilitate training, the multi-path factorized convolutions allow a significant reduction of the convolutional layers. On the other hand, in contrast to previous lightweight networks \cite{Badrinarayanan2015Segnet,Paszke2016enet,Romera2018erfnet} that abstract feature representation with fixed filter kernel size, the FCUs adopt the 1D factorized convolutions with different kernel size, where the receptive fields are adaptive to capture object instances with different scales. The FCUs and PFCUs are symmetrically stacked to construct our encoder-decoder architecture, producing semantic segmentation output end-to-end in the same resolution as the input image. Although the focus of this paper is the task of semantic segmentation, the proposed FCUs and PFCUs is directly transferable to any existing network that makes use of residual layers, including both classification and segmentation architectures. In summary, our contributions are three-folds: (1) The symmetrical architecture of ESNet leads to the great reduction of network complexity, accelerating the entire inference process; (2) Using multiple branch parallel convolutions in the residual layer leverages network size and powerful feature representation, still resulting in the whole network can be trained end-to-end. (3) We evaluate the performance of ESNet on CityScapes dataset \cite{Cordts2016the}, and the experimental results show that compared with recent mainstream lightweight networks, it achieves the best available trade-off in terms of accuracy and efficiency.

The remainder of this paper is organized as follows. After a brief discussion of related work in Section \ref{sec:RleatedWork}, a fast and compact architecture named ESNet is proposed in Section \ref{sec:ESNet}. The proposed network has been evaluated on CityScapes dataset, where the benchmark is constructed on a single NVIDIA GTX 1080Ti GPU. These experiments can be found in Section \ref{sec:Experiments}. Finally, the conclusion  remarks and future work are given in Section \ref{sec:Conclusion}.

\section{Related Work}\label{sec:RleatedWork}

Recent success in DCNNs has brought significant progress on semantic segmentation in the past few years \cite{girshick2014rich,long2017fully,Chen2016deeplab,zhao2017pyramid}. As a pioneer work, Farabet et. al \cite{farabet2013learning} utilized DCNNs to abstract hierarchical feature representation for semantic segmentation. In \cite{long2017fully}, Long et. al first proposed an end-to-end segmentation based on VGG-16 network, where the fully connected layers in traditional DCNNs are replaced by convolutional layers to upsample feature maps. So far, a large amount of FCN-based networks \cite{Chen2016deeplab,zhao2017pyramid,xiao2017not,understand2018wang,Chen2017rethinking} have been designed to deal with semantic segmentation challenge. To enlarge receptive fields, the dilated convolution \cite{yu2015multi} or atrous convolution \cite{Chen2016deeplab,Chen2017rethinking} is also employed in FCN to capture large scale context. Due to continuous pooling, however, the resolution of feature maps are significant reduced and directly adapting FCNs always leads to the poor estimation outputs. To refine predict results, the encoder-decoder networks \cite{Badrinarayanan2015Segnet,Guosheng2017RefineNet,noh2015learning,chao2017large} are commonly-used to develop FCN architecture by sequentially recovering fine image details. For instance, Noh et al. employ deconvolution to upsample low resolution feature responses \cite{noh2015learning}. SegNet \cite{Badrinarayanan2015Segnet} reuses the recorded pooling indices to upsample feature maps, and learns extra deconvolutional layers to densify the feature responses. Through adding skip connections, U-Net \cite{olaf2015unet} designs an elegant symmetric network architecture, which stacks convolutional features from the encoder to the decoder activations. More recently, more attention have been paid to RefineNets \cite{Guosheng2017RefineNet,chao2017large,tobias2017full,md2017gate}, which adopt ResNet \cite{he2016deep} in encoder-decoder structure, and have been demonstrated very effective on several semantic segmentation benchmarks \cite{Cordts2016the,everingham2015pascal}.

In spite of achieving promising performance, these advances are at the sacrifice of running time and speed. In order to overcome this problem, many lightweight networks, initially designed for image classification task \cite{Howard2017mobile,Rastegari2016xnor,zhang2018shuffle,Wu2016quantized,xie2017agg}, have been designed to balance the segmentation accuracy and implementing efficiency \cite{wu2018cgnet,Treml2016speeding,Paszke2016enet,Mehta2018espnet,yu2018bisenet,Zhao2018ICnet}. ENet \cite{Paszke2016enet} is the first work that considers the efficiency issue, where the point-wise convolution is adopted in the residual layer. Apart from this initial designment, some recent work always employ convolution factorization principle \cite{szegedy2015going,Howard2017mobile,Szegedy2016rethinking} in their network architecture, where the 2D standard convolution is replaced by depthwise separable convolution. For example, Zhao et al. \cite{Zhao2018ICnet} investigate the high-level label cues to improve performance. ERFNet \cite{Romera2018erfnet} leverages skip connections and 1D convolutions in residual block designment, greatly reducing network parameters while maintaining high efficiency. In \cite{Mehta2018espnet}, Mehta et al. design an efficient spatial pyramid convolution network for semantic segmentation. Some similar networks also use symmetrical encoder-decoder architecture \cite{Badrinarayanan2015Segnet,Mehta2018espnet,Romera2018erfnet,fast2019zhang}, while the other approaches take the contextual clues into account \cite{wu2018cgnet,yu2018bisenet} to balance performance and efficiency. Unlike these lightweight networks, our ESNet utilizes multiple branch parallel factorized convolution, achieving real-time inference and higher accuracy.

\section{ESNet}\label{sec:ESNet}

In this section, we first introduce the whole architecture of ESNet, and then elaborate on the designed details of each unit.

\subsection{Network Overview}\label{sec:Network}

\begin{table}[!t]
\tabcolsep 0.6mm \caption{The architecture of ESNet. ``Size'' denotes the dimension of output feature maps, $C$ is the number of classes.}
\begin{center}
\begin{tabular}{|c||c|c|c|c|}
\hline
\textbf{Stage} & \textbf{Name} & \textbf{Layer} & \textbf{Type} &\textbf{Size} \\
\hline
\hline
\multirow{8}*{\rotatebox{90} {Encoder}} &\multirow{2}*{Block 1}
&1  &\textbf{Down-sampling Unit}     &$512 \times 256 \times 16$ \\
&~  &2-4 &$3 \times$ \textbf{FCU} ($K = 3$) &$512 \times 256 \times 16$ \\ \cline{2-5}
&\multirow{2}*{Block 2}
&5  &\textbf{Down-sampling Unit}     &$256 \times 128 \times 64$ \\
&~  &6-7 &$2 \times$ \textbf{FCU} ($K = 5$) &$256 \times 128 \times 64$ \\ \cline{2-5}
&\multirow{2}*{Block 3}
&8  &\textbf{Down-sampling Unit}     &$128 \times 64 \times 128$ \\
&~  &9-11  &$3 \times$ \textbf{PFCU} (dilated $r_1 = 2, r_2 = 5, r_3 = 9$)  &$128 \times 64 \times 128$ \\
\hline
\multirow{5}*{\rotatebox{90} {Decoder}}
&\multirow{2}*{Block 4}
&12 &\textbf{Up-sampling Unit}     &$256 \times 128 \times 64$ \\
&~  &13-14  &$2 \times$ \textbf{FCU} ($K = 5$) &$256 \times 128 \times 64$ \\ \cline{2-5}
&\multirow{2}*{Block 5}
&15 &\textbf{Up-sampling Unit}     &$512 \times 256 \times 16$ \\
&~  &15-17  &$2 \times$ \textbf{FCU} ($K = 3$) &$512 \times 256 \times 16$ \\ \cline{2-5}
&\multirow{1}*{Full Conv}
&18 &\textbf{Up-sampling Unit}     &$1024 \times 512 \times C$ \\
\hline
\end{tabular}
\end{center}\label{tab:ESNet}
\end{table}

As shown in Table \ref{tab:ESNet} and illustrated in Figure \ref{fig:Overview}, our ESNet has a symmetric encoder-decoder architecture, where an encoder produces downsampled feature maps, and a decoder upsamples the feature maps to match input resolution. The entire network is composed of 18 convolution layers, where the residual module is adopted as our core element. As shown in Table \ref{tab:ESNet}, the encoder and decoder has nearly same number of convolution layers, and utilize similar convolution type. For instance, both Block 1 and Block 5 employ FCU with $K =3$, while Block 2 and Block 4 also employ FCU with $K =5$. As illustrated in Figure \ref{fig:Overview}, the input image first undergoes a down-sampling unit to form initial feature maps, which are fed into the subsequent residual layers. Downsampling enables more deeper network to gather context, while at the same time helps to reduce computation. Additionally, two types of residual convolution module, called FCU and PFCU are employed, where the first one uses factorized convolution to extract low-level features, and the second one utilizes multi-branch dilated convolution to enlarge receptive fields to capture high-level semantics. In \cite{Paszke2016enet,Romera2018erfnet,fast2019zhang}, the designed networks are began with sustained downsampling, however, such kind of operation may be harmful to feature abstraction, which highly detriment segmentation accuracy. In order to address this problem, our ESNet postpones downsampling oepration in encoder, with the similar spirit of \cite{Szegedy2016rethinking}. In the following, we will describe how to design FCU and PFCU, which focus on solving the efficiency limitation that is essentially present in the residual layer.

\subsection{FCU Module}\label{sec:FCU}

\begin{figure}[!t]
\centerline{\includegraphics[width = 1.0\textwidth]{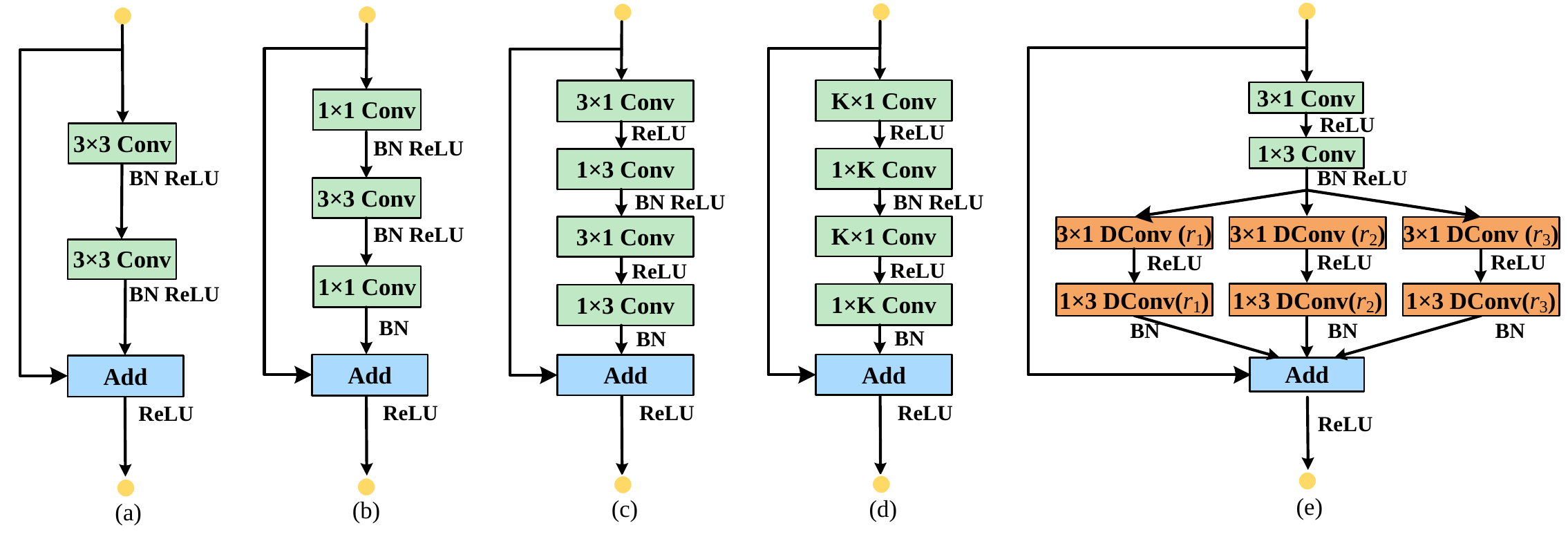}}
\caption{Comparison of different residual layer modules. From left to right are (a) Non-bottleneck \cite{he2016deep}, (b) Bottleneck \cite{Paszke2016enet}, (c) Non-bottleneck-1D \cite{Romera2018erfnet}, (d) FCU and (e) PFCU module. ``DConv'' denotes the dilated convolution, where $r_1$, $r_2$, and $r_3$ are dilated rates for each split branch, respectively.} \label{fig:ResidualModule}
\end{figure}

To reduce computation budget, the pointwise convolutions \cite{Paszke2016enet,zhang2018shuffle,xie2017agg} and factorized convolutions \cite{Romera2018erfnet,fast2019zhang} are widely used to take place of traditional standard convolution in residual layer. Essentially, pointwise convolution (e.g., $1 \times 1$) speeds up computation by reducing the number of convolutional channels, which thus can be also considered as dimensionality reduction of feature maps. On the other hand, factorized convolution attempts to perform convolution with smaller filter kernel size. As a result, the recent years have witnessed multiple successful instances of lightweight residual layer \cite{Paszke2016enet,Howard2017mobile}, such as Non-bottleneck (Figure \ref{fig:ResidualModule} (a)), Bottleneck (Figure \ref{fig:ResidualModule} (b)), and Non-bottleneck-1D (Figure \ref{fig:ResidualModule} (c)). More specifically, the Bottleneck module comes from the standard residual layer of ResNet \cite{he2016deep}, which requires less computational resources with respect to Non-bottleneck module. Although it is commonly adopted in state-of-the-art networks \cite{Paszke2016enet,Mehta2018espnet,Howard2017mobile}, the performance descend drastically when network goes deeper. Another outstanding residual module is Non-bottleneck-1D \cite{Romera2018erfnet}, which can be considered as a special case of our FCU with $K = 3$. In this module, a standard $3 \times 3$ convolution in the bottleneck is decomposed into two 1D convolutions (e.g., $3 \times 1$ and $1 \times 3$), yet the fixed kernel size of factorized convolution limits the field-of-view, leading to the decrease of performance.

As shown in Figure \ref{fig:ResidualModule} (d), we again employ Non-bottleneck-1D module \cite{Romera2018erfnet,fast2019zhang} to design our FCU, as this is helpful for greatly reducing the number of parameters and accelerating the training and inference process. Unlike previous approaches \cite{Romera2018erfnet,fast2019zhang}, however, the size of convolutional kernel is unfixed, allowing FCU to adaptively broaden receptive fields. For example, in the encoder, the shallow layers (Block 1 in Table \ref{tab:ESNet}) prefer to use smaller kernel size ($K = 3$) to abstract low-level image features, while deeper layers (Block 2 in Table \ref{tab:ESNet}) resort to larger kernel size ($K = 5$) for capturing wide-scale context. Conversely, in the decoder, the shallow layers (Block 4 in Table \ref{tab:ESNet}) utilize a larger kernel size ($K = 5$) to gather long-ranged information to enhance prediction accuracy, while deeper layers (Block 5 in Table \ref{tab:ESNet}) symmetrically employ smaller kernel size ($K = 3$) to recover image details by smoothing filter responses of short-ranged neighborhood pixels.

\subsection{PFCU Module}\label{sec:PFCU}

We focus on solving the accuracy and efficiency trade-off as a whole, without sitting on only one of its sides. To this end, this section introduces PFCU, as depicted in  Figure \ref{fig:ResidualModule}(e). Motivated from \cite{Rastegari2016xnor,Szegedy2016rethinking}, a \emph{transform-split-transform-merge} strategy is employed in the designment of our PFCU, where each branch employs dilated convolution with different rate to broaden receptive fields. The dilated convolutions with parallel multiple branch are adaptive to capture objects within different scales, approaching the representational power of large and dense layers, but at a considerably lower computational complexity. At the beginning of each PFCU, the input is first transferred by a set of specialized 1D filters (e.g., $1 \times 3$ and $3 \times 1$), and the convolutional outputs pass through three parallel dilated convolution with rates $r_1 = 2$, $r_2 = 5$, and $r_3 = 9$, respectively. To facilitate training, finally, the outputs of three convolutional branches are added with input through the branch of identity mapping. After merging, the next PFCU begins. It is clear that our PFCU is not only efficient, but also accurate. Firstly, the powerful representation ability of PFCU allows us to use less convolution layers. Secondly, in PFCU, each branch shares the same convolutional feature maps. This can be regarded as a kind of feature reuse, which to some extent enlarges network capacity without significantly increasing complexity.

\begin{table}[!t]
\tabcolsep 0.3mm \caption{Comparison of weights used in different type of residue blocks. Three parameters of ``Size'' are number of layers, feature channels and convolutional kernels in corresponding residue block, respectively.}
\begin{center}
\begin{tabular}{|c|c||c|c|c|c|c|c|}
\hline
\multicolumn{2}{|c||}{Method}     & \multicolumn{3}{c|}{Encoder}     & \multicolumn{2}{c|}{Decoder} & \#Para                 \\ \hline \hline
\multirow{2}{*}{ERFNet\cite{Romera2018erfnet}} & Type   & Non-bt-1D & -        & Non-bt-1D & Non-bt-1D     & Non-bt-1D    & \multirow{2}{*}{17,688} \\ \cline{2-7}
                        & Size & $5\times64\times12$   & -        & $8\times128\times12$  & $2\times64\times12$       & $2\times16\times12$      &                        \\ \hline
\multirow{2}{*}{ESNet}   & Type   & FCU(K=3)  & FCU(K=5) & PFCU      & FCU(K=5)      & FCU(K=3)     & \multirow{2}{*}{15,296} \\ \cline{2-7}
                        & Size & $3\times16\times12$   & $2\times64\times20$  & $3\times128\times24$  & $2\times64\times20$       & $2\times16\times12$      &                        \\ \hline
\end{tabular}
\end{center}\label{tab:Complexity}
\end{table}

\subsection{Comparison of Network Complexity}

In this section, we analyze the network complexity of our ESNet and compare with recent state-of-the-art ERFNet \cite{Romera2018erfnet}. In addition, we also compare our new implementation of the residual layer that makes use of the parallel 1D factorization to accelerate and reduce the parameters. Table \ref{tab:Complexity} summarizes the total dimensions of the weights on the convolutions of every residual block. As shown in Figure \ref{fig:ResidualModule}, Non-Bottleneck-1D has the simplest structure and fewest parameters. Since the standard $3 \times 3$ convolution has been decomposed into $1 \times 3$ and $3 \times 1$ convolution, the number of convolutional kernels is only 12 in each residual layer. As our FCU (K =3) module also adopts 1D factorized convolution, it has the same number of kernels with respect to Non-Bottleneck-1D. On the other hand, FCU (K = 5) and PFCU involve larger kernel size or more factorized convolution, leading to the increase of convolutional kernels (20 for FCU (K = 5), and 24 for PFCU) used in corresponding residual layers. However, the total weights are not only decided by filter kernel size, but also depend on the number of feature channels and convolution layers. Due to the parallel design of PFCU that facilitates the reduction of convolution layers, the entire size of ESNet is still smaller than ERFNet \cite{Romera2018erfnet}. For example, in contrast to ERFNet \cite{Romera2018erfnet} that contains 8 layers of dilated convolution ($8 \times 128 \times 12 = 12,288$), our ESNet only has 3 layers of PFCU, resulting in more fewer parameters ($3 \times 128 \times 24 = 9,216$) while achieving higher accuracy. As for the total parameters, our ESNet design is clearly more benefited, by receiving a direct 13.5\% reduction, and thus greatly accelerates its execution. This is also consistent with the results of Table \ref{tab:Result1}.

\section{Experiments}\label{sec:Experiments}

In this section, we carry on the experiments to demonstrate the potential of our segmentation architecture in terms of accuracy and efficiency trade-off.

\subsection{Dataset}

The widely-used CityScapes dataset \cite{Cordts2016the}, including 19 object classes and one additional background, is selected to evaluate our ESNet. Beside the images with fine pixel-level annotations that contain 2,975 training, 500 validation and 1,525 testing images, we also use the 20K coarsely annotated images for training.

\subsection{Implementation Details}

To show the advantages of ESNet, we selected 6 state-of-the-art lightweight networks as baselines, including SegNet \cite{Badrinarayanan2015Segnet}, ENet \cite{Paszke2016enet}, ERFNet \cite{Romera2018erfnet}, ICNet \cite{Zhao2018ICnet}, CGNet \cite{wu2018cgnet}, and ESPNet \cite{Mehta2018espnet}. We adopt mean intersection-over-union (mIOU) averaged across all classes and categories to evaluate segmentation accuracy, while running time, inference speed (FPS), and model size (number of parameters) to measure implementing efficiency. For fair comparison, all the methods are conducted on the same hardware platform of DELL workstation with a single GTX 1080Ti GPU. We favor a large minibatch size (set as 4) to make full use of the GPU memory, where the initial learning rate is $5 \times 10^{-4}$ and the `poly' learning rate policy is adopted with power 0.9, together with momentum and weight decay are set to 0.9 and $10^{-4}$, respectively.

\subsection{Evaluation Results}

\begin{table}[!t]
\tabcolsep 3.2mm \caption{Comparison with the state-of-the-art approaches in terms of segmentation accuracy and implementing efficiency.}
\begin{center}
\begin{tabular}{|c||ccccc|}
\hline
Method & Cla(\%) &Cat(\%) &Time(ms) &Speed(Fps) &Para(M) \\
\hline
\hline
SegNet\cite{Badrinarayanan2015Segnet}  &57.0 &79.1 &67 &15 &29.5\\
ENet\cite{Paszke2016enet}              &58.3 &80.4 &13 &\textbf{77} &\textbf{0.36}\\
ESPNet\cite{Mehta2018espnet}           &60.3 &82.2 &18 &54 &0.40\\
CGNet\cite{wu2018cgnet}                &64.8 &85.7 &20 &50 &0.50\\
ERFNet \cite{Romera2018erfnet}         &66.3 &86.5 &21 &48 &2.10\\
ICNet \cite{Zhao2018ICnet}             &69.5 &86.4 &33 &30 &7.80\\
\hline
Ours                                   &\textbf{70.7} &\textbf{87.4} &16 &63 &1.66\\
\hline
\end{tabular}
\end{center}\label{tab:Result1}
\end{table}

\begin{table}[!t]
\tabcolsep 1.25mm \caption{Individual category results on the CityScapes test set in terms of class and category mIOU scores. Methods trained using both fine and coarse data are marked with superscript `${\dag}$'. The best performance for each individual class is marked with bold-face number.}
\begin{center}
\begin{tabular}{|c|cccccccccc|c|}
\hline
Method  &{Roa}  &{Sid}  &{Bui}  &{Wal}  &{Fen}  &{Pol}  &{TLi}  &{TSi}  &{Veg}  &{Ter}    &{Cla}\\
\hline
\hline
SegNet \cite{Badrinarayanan2015Segnet} &96.4 &73.2 &84.0 &28.4 &29.0 &35.7 &39.8 &45.1 &87.0 &63.8 &57.0\\
ENet \cite{Paszke2016enet}             &96.3 &74.2 &75.0 &32.2 &33.2 &43.4 &34.1 &44.0 &88.6 &61.4 &58.3\\
ESPNet \cite{Mehta2018espnet}          &97.0 &77.5 &76.2 &35.0 &36.1 &45.0 &35.6 &46.3 &90.8 &63.2 &60.3\\
CGNet \cite{wu2018cgnet}               &95.5 &78.7 &88.1 &40.0 &43.0 &54.1 &59.8 &63.9 &89.6 &67.6 &64.8\\
ERFNet \cite{Romera2018erfnet}         &97.2 &80.0 &89.5 &41.6 &45.3 &56.4 &60.5 &64.6 &91.4 &\textbf{68.7} &66.3\\
ICNet \cite{Zhao2018ICnet}             &97.1 &79.2 &89.7 &43.2 &48.9 &\textbf{61.5} &60.4 &63.4 &91.5 &68.3 &69.5\\
\hline
Ours                                   &97.1 &78.5 &90.4 &46.5 &48.1 &60.1 &60.4 &70.9 &91.1 &59.9 &69.1\\
Ours$^{\dag}$                          &\textbf{98.1} &\textbf{80.4} &\textbf{92.4} &\textbf{48.3} &\textbf{49.2} &\textbf{61.5} &\textbf{62.5} &\textbf{72.3} &\textbf{92.5} &61.5 &\textbf{70.7}\\
\hline
\hline
Method  &{Sky}  &{Ped}  &{Rid}  &{Car}  &{Tru}  &{Bus}  &{Tra}  &{Mot}  &{Bic}  &{~}    &{Cat}\\
\hline
\hline
SegNet \cite{Badrinarayanan2015Segnet} &91.8 &62.8 &42.8 &89.3 &38.1 &43.1 &44.1 &35.8 &51.9 &~ &79.1\\
ENet \cite{Paszke2016enet}             &90.6 &65.5 &38.4 &90.6 &36.9 &50.5 &48.1 &38.8 &55.4 &~ &80.4\\
ESPNet \cite{Mehta2018espnet}          &92.6 &67.0 &40.9 &92.3 &38.1 &52.5 &50.1 &41.8 &57.2 &~ &82.2\\
CGNet \cite{wu2018cgnet}               &92.9 &74.9 &54.9 &90.2 &44.1 &59.5 &25.2 &47.3 &60.2 &~ &85.7\\
ERFNet \cite{Romera2018erfnet}         &94.2 &76.1 &\textbf{56.4} &92.4 &45.7 &60.6 &27.0 &48.7 &61.8 &~ &86.5\\
ICNet \cite{Zhao2018ICnet}             &93.5 &74.6 &56.1 &92.6 &51.3 &72.7 &51.3 &\textbf{53.6} &70.5 &~ &86.4\\
\hline
Ours                                   &93.2 &74.3 &51.8 &92.3 &61.0 &72.3 &51.0 &43.3 &70.2 &~ &86.8\\
Ours$^{\dag}$                          &\textbf{94.4} &\textbf{76.6} &53.2 &\textbf{94.4} &\textbf{62.5} &\textbf{74.3} &\textbf{52.4} &45.5 &\textbf{71.4} &~ &\textbf{87.4}\\
\hline
\end{tabular}
\end{center}\label{tab:Result2}
\end{table}

\begin{figure*}[!t]
\centerline{\includegraphics[width = 1.0\textwidth]{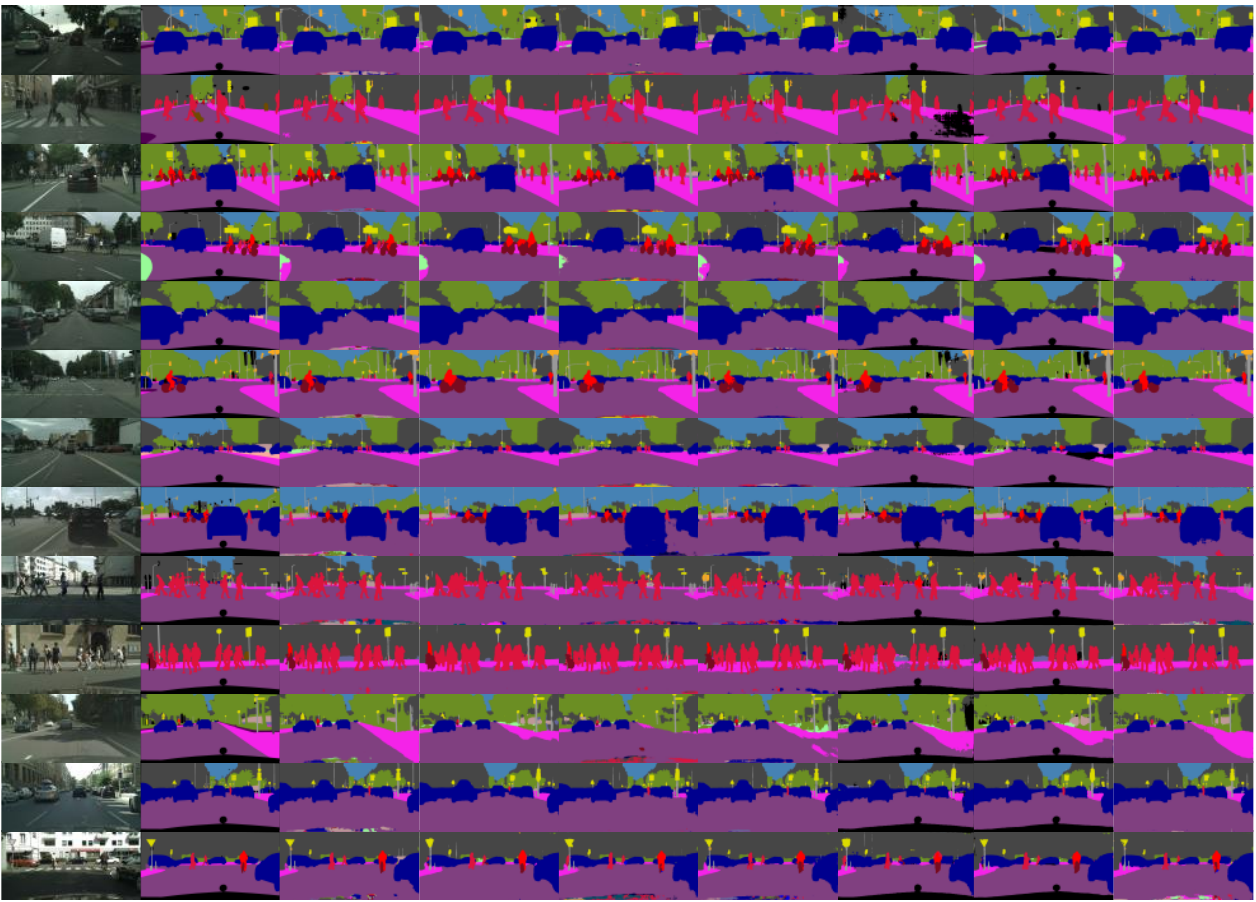}}
\caption{The visual comparison on CityScapes val dataset. From left to right are input images, ground truth, segmentation outputs from our ESNet, SegNet \cite{Badrinarayanan2015Segnet}, ENet \cite{Paszke2016enet}, ERFNet \cite{Romera2018erfnet}, ESPNet \cite{Mehta2018espnet}, ICNet \cite{Zhao2018ICnet}, and CGNet \cite{wu2018cgnet}. (Best viewed in color)} \label{fig:Result3}
\end{figure*}

In Table \ref{tab:Result1} and Table \ref{tab:Result2}, we have reported the quantitative results compared with state-of-the-art baselines. The results demonstrate that ESNet achieves the best available trade-off in terms of accuracy and efficiency. Without data augmentation, our ESNet obtains comparable results with respect to ICNet \cite{Zhao2018ICnet} (only slight 0.4\% drop of class mIOU, but 0.4\% improvement of category mIOU). After augmented with 20K additional data with coarse annotations, our ESNet yields 70.7\% class mIOU and 87.4\% category mIOU, respectively, where 16 out of the 19 categories obtains best scores. Regarding to the efficiency, ESNet is nearly $4 \times$ faster and $18 \times$ smaller than SegNet \cite{Badrinarayanan2015Segnet}. Although ENet \cite{Paszke2016enet}, an anther efficient network, is nearly $1.2 \times$ efficient, and has $5 \times$ less parameters than our ESNet, but delivers poor segmentation accuracy of 12.4\% and 7\% drops in terms of class and category mIOU, respectively. Another interesting results is the comparison with CGNet \cite{wu2018cgnet} in Table \ref{tab:Result1}, where it has $3 \times$ fewer parameters, while performs slightly slower than our ESNet. This is probably because that ESNet has more simpler architecture and less convolution layers, yielding more efficient in inference process. Figure \ref{fig:Result3} shows some visual examples of segmentation outputs on the CityScapes validation set. It is demonstrated that, compared with baselines, our ESNet not only correctly classifies object with different scales (especially for very small object instance, such as ``traffic sign'' and ``traffic light''), but also produces consistent qualitative results for all classes.

\section{Conclusion Remark and Future Work}\label{sec:Conclusion}

This paper has proposed an architecture that achieves accurate and fast pixel-wise semantic segmentation. In contrast to top-accurate networks that are computationally expensive with complex and deep architectures, our ESNet focuses more on developing the core elements of network architecture: the convolutional blocks. The transform-split-transform-merge scheme is adopted to redesign the commonly-used residual layers, leading to the multi-branch parallel 1D decomposed convolution, which is more efficient while retaining a similar learning performance. As this design can be directly used in existing encoder-decoder networks, we propose an ESNet that completely leverages its benefits to reach state-of-the-art segmentation accuracy and efficiency. The experimental results show that our ESNet achieves best available trade-off on CityScapes dataset in terms of segmentation accuracy and implementing efficiency. The future work includes incorporating contextual branch, as well as \cite{wu2018cgnet,yu2018bisenet} does, to further improve performance while remaining few parameters.



%
%
%
%

\bibliographystyle{splncs}
\bibliography{egbib}

\begin{thebibliography}{10}

\bibitem{krizhevsky2012imagenet}
Krizhevsky, A., Sutskever, I., Hinton, G.E.:
\newblock Imagenet classification with deep convolutional neural networks.
\newblock In: NIPS. (2012)  1097--1105

\bibitem{he2016deep}
He, K., Zhang, X., Ren, S., Sun, J.:
\newblock Deep residual learning for image recognition.
\newblock In: CVPR. (2016)  770--778

\bibitem{girshick2014rich}
Girshick, R., Donahue, J., Darrell, T., Malik, J.:
\newblock Rich feature hierarchies for accurate object detection and semantic
  segmentation.
\newblock In: CVPR. (2014)  580--587

\bibitem{long2017fully}
Long, J., Shelhamer, E., Darrell, T.:
\newblock Fully convolutional networks for semantic segmentation.
\newblock IEEE TPAMI \textbf{39} (2017)  640--651

\bibitem{Chen2016deeplab}
Chen, L.C., Papandreou, G., Kokkinos, I., Murphy, K., Yuille, A.L.:
\newblock Deeplab: Semantic image segmentation with deep convolutional nets,
  atrous convolution, and fully connected crfs.
\newblock IEEE TPAMI \textbf{40} (2018)  834--848

\bibitem{zhao2017pyramid}
Zhao, H., Shi, J., Qi, X., Wang, X., Jia, J.Y.:
\newblock Pyramid scene parsing network.
\newblock In: CVPR. (2016)  6230--6239

\bibitem{xiao2017not}
Xiaoxiao, L., Zhiwei, L., Ping, L., Chenchange, L., Xiaoou, T.:
\newblock Not all pixels are equal: Difficulty-aware semantic segmentation via
  deep layer cascade.
\newblock In: CVPR. (2017)  6459--6468

\bibitem{Badrinarayanan2015Segnet}
Badrinarayanan, V., Alex, K., Roberto, C.:
\newblock Segnet: A deep convolutional encoder-decoder architecture for image
  segmentation.
\newblock arXiv preprint arXiv:1511.00561 (2015)

\bibitem{Guosheng2017RefineNet}
Guosheng, L., Anton, M., Chunhua, S., Reid, I.:
\newblock Refinenet: multi-path refinement networks for high-resolution
  semantic segmentation.
\newblock In: CVPR. (2017)  5168--5177

\bibitem{noh2015learning}
Noh, H., Hong, S., Han, B.:
\newblock Learning deconvolution network for semantic segmentation.
\newblock In: ICCV. (2015)  1520--1528

\bibitem{szegedy2015going}
Szegedy, C., Liu, W., Jia, Y., Sermanet, P., Reed, S., Anguelov, D., Erhan, D.,
  Vanhoucke, V., Rabinovich, A.:
\newblock Going deeper with convolutions.
\newblock In: CVPR. (2015)  1--9

\bibitem{chao2017large}
Peng, C., Xiangyu, Z., Gang, Y., Guiming, L., Jian, S.:
\newblock Large kernel matters: Improve semantic segmentation by global
  convolutional network.
\newblock In: CVPR. (2017)  1743--1751

\bibitem{exploring2018lin}
Lin, G.S., Shen, C.H., Van, D.H., Reid, I.:
\newblock Exploring context with deep structured models for semantic
  segmentation.
\newblock IEEE TPAMI \textbf{40} (2018)  1352--1366

\bibitem{cong2019can}
Cong, D., Zhou, Q., Chen, J., Wu, X., Zhang, S., Ou, W., Lu, H.:
\newblock Can: Contextual aggregating network for semantic segmentation.
\newblock In: ICASSP. (2019)  accepted

\bibitem{wu2018cgnet}
Wu, T.Y., Tang, S., Zhang, R., Zhang, Y.D.:
\newblock Cgnet: A light-weight context guided network for semantic
  segmentation.
\newblock In: arXiv preprint arXiv:1811.08201v1. (2018)

\bibitem{Treml2016speeding}
Treml, M., Arjona-Medina, J., Mayr, A., Heusel, M., Widrich, M., Bodenhofer,
  U., Nessler, B., Hochreiter, S.:
\newblock Speeding up semantic segmentation for autonomous driving.
\newblock In: NIPS Workshop. (2016)  1--7

\bibitem{Paszke2016enet}
Paszke, A., Chaurasia, A., Kim, S., Culurciello, E.:
\newblock Enet: A deep neural network architecture for real-time semantic
  segmentation.
\newblock In: arXiv preprint arXiv:1606.02147. (2016)

\bibitem{Mehta2018espnet}
Mehta, S., Rastegari, M., Caspi, A., Shapiro, L., Hajishirzi, H.:
\newblock Espnet: Efficient spatial pyramid of dilated convolutions for
  semantic segmentation.
\newblock In: arXiv preprint arXiv:1803.06815v3. (2018)

\bibitem{Romera2018erfnet}
Romera, E., Alvarez, J.M., Bergasa, L.M., Arroyo, R.:
\newblock Erfnet: Efficient residual factorized convnet for real-time semantic
  segmentation.
\newblock IEEE TITS \textbf{19} (2018)  263--272

\bibitem{Cordts2016the}
Cordts, M., Omran, M., Ramos, S., Rehfeld, T., Enzweiler, M., Benenson, R.,
  Franke, U., Roth, S., Schiele, B.:
\newblock The cityscapes dataset for semantic urban scene understanding.
\newblock In: CVPR. (2016)  3213--3223

\bibitem{farabet2013learning}
Farabet, C., Couprie, C., Najman, L., LeCun, Y.:
\newblock Learning hierarchical features for scene labeling.
\newblock IEEE TPAMI \textbf{35} (2013)  1915--1929

\bibitem{understand2018wang}
Panqu, W., Pengfei, C., Ye, Y., Ding, L., Zehua, H., Xiaodi, H., Cottrell, G.:
\newblock Understanding convolution for semantic segmentation.
\newblock In: WACV. (2018)  1451--1460

\bibitem{Chen2017rethinking}
Liang-Chieh, C., George, P., F., S., H., A.:
\newblock Rethinking atrous convolution for semantic image segmentation.
\newblock In: arXiv:1706.05587. (2017)

\bibitem{yu2015multi}
Yu, F., Koltun, V.:
\newblock Multi-scale context aggregation by dilated convolutions.
\newblock arXiv preprint arXiv:1511.07122 (2015)

\bibitem{olaf2015unet}
Ronneberger, O., Philipp, F., Thomas, B.:
\newblock U-net: Convolutional networks for biomedical image segmentation.
\newblock In: MICCAI. (2015)  225--233

\bibitem{tobias2017full}
Pohlen, T., Hermans, A., Mathias, M., Leibe, B.:
\newblock Full-resolution residual networks for semantic segmentation in street
  scenes.
\newblock In: CVPR. (2017)  3309--3318

\bibitem{md2017gate}
Islam, M.A., Rochan, M., Bruce, N.D.B., Wang, Y.:
\newblock Gated feedback refinement network for dense image labeling.
\newblock In: CVPR. (2017)  4877--4885

\bibitem{everingham2015pascal}
Everingham, M., Eslami, S.A., Van~Gool, L., Williams, C.K., Winn, J.,
  Zisserman, A.:
\newblock The pascal visual object classes challenge: A retrospective.
\newblock IJCV \textbf{111} (2015)  98--136

\bibitem{Howard2017mobile}
Howard, A.G., Zhu, M., Chen, B., Kalenichenko, D., W.Wang, Weyand, T.,
  Andreetto, M., Adam, H.:
\newblock Mobilenets: efficient convolutional neural networks for mobile vision
  applications.
\newblock In: arXiv preprint arXiv:1704.04861. (2017)

\bibitem{Rastegari2016xnor}
Rastegari, M., Ordonez, V., Redmon, J., Farhadi, A.:
\newblock Xnor-net: Imagenet classification using binary convolutional neural
  networks.
\newblock In: ECCV. (2016)

\bibitem{zhang2018shuffle}
Zhang, X., Zhou, X., Lin, M., Sun, J.:
\newblock Shufflenet: An extremely efficient convolutional neural network for
  mobile devices.
\newblock In: CVPR. (2018)  6848--6856

\bibitem{Wu2016quantized}
Wu, J., Leng, C., Wang, Y., Hu, Q., Cheng, J.:
\newblock Quantized convolutional neural networks for mobile devices.
\newblock In: CVPR. (2016)  5168--5177

\bibitem{xie2017agg}
Xie, X., Girshick, R., Dollar, P., Tu, Z.W., He, K.M.:
\newblock Aggregated residual transformations for deep neural networks.
\newblock In: CVPR. (2017)  5987--5995

\bibitem{yu2018bisenet}
Changqian, Y., Jingbo, W., Chao, P., Changxin, G., Gang, Y., Nong, S.:
\newblock Bisenet: Bilateral segmentation network for real-time semantic
  segmentation.
\newblock In: arXiv preprint arXiv:1808.00897. (2018)

\bibitem{Zhao2018ICnet}
Zhao, H.S., Qi, X.J., Shen, X.Y., Shi, J.P., Jia, J.Y.:
\newblock Icnet for real-time semantic segmentation on high-resolution images.
\newblock In: arXiv preprint arXiv:1704.08545v2. (2018)

\bibitem{Szegedy2016rethinking}
Szegedy, C., Vanhoucke, V., Ioffe, S., Shlens, J., Wojna, Z.:
\newblock Rethinking the inception architecture for computer vision.
\newblock In: CVPR. (2016)  2818--2826

\bibitem{fast2019zhang}
Zhang, X., Cheny, Z., Wu, Q.M.J., Cai, L., Lu, D., Li, X.:
\newblock Fast semantic segmentation for scene perception.
\newblock IEEE TII (2019)  accepted

\end{thebibliography}

\end{document}